\documentclass{article}

\usepackage[final]{nips_2016}

\usepackage[utf8]{inputenc} 
\usepackage[T1]{fontenc}    
\usepackage{hyperref}       
\usepackage{url}            
\usepackage{booktabs}       
\usepackage{amsfonts}       
\usepackage{nicefrac}       
\usepackage{microtype}      
\usepackage{amsmath}
\usepackage{amssymb}
\usepackage{bm}
\usepackage{hyperref}
\usepackage{placeins}
\usepackage{caption}
\usepackage{subcaption}
\usepackage{tikz}
\usepackage{tikzsymbols}
\usepackage{todonotes}
\usepackage{bbm}
\usepackage{color}

\usepackage{graphicx}
\usepackage{float} 
\usepackage{animate}
\usetikzlibrary{positioning,calc}
\usepackage[most]{tcolorbox}
\usepackage{enumerate}
\usepackage{fancyvrb}
\usepackage{titlesec}
\titlespacing\section{0pt}{12pt plus 0pt minus 0pt}{0pt plus 0pt minus 0pt}
\titlespacing\subsection{0pt}{12pt plus 0pt minus 0pt}{0pt plus 0pt minus 0pt}
\titlespacing\subsubsection{0pt}{12pt plus 0pt minus 0pt}{0pt plus 0pt minus 0pt}

\title{Grounding Complex Navigational Instructions Using Scene Graphs}

\author{
    Michiel de Jong* \\
    Carnegie Mellon University \\
   \texttt{mdejong@andrew.cmu.edu}
   \And
	Satyapriya Krishna* \\
	Carnegie Mellon University \\
   \texttt{satyaprk@andrew.cmu.edu}
   \And
  Anuva Agarwal\\
  Carnegie Mellon University \\
   \texttt{anuvaa@andrew.cmu.edu}
}

\begin{document}
\maketitle{}

\begin{abstract}
Training a reinforcement learning agent to carry out natural language instructions is limited by the available supervision, i.e. knowing when the instruction has been carried out. We adapt the CLEVR visual question answering dataset to generate complex natural language navigation instructions and accompanying scene graphs, yielding an environment-agnostic supervised dataset. To demonstrate the use of this data set, we map the scenes to the VizDoom environment and use the architecture in \citet{gatedattention} to train an agent to carry out these more complex language instructions.
\end{abstract} 

\section{Introduction}
When humans assign tasks, they provide instructions in natural language. An artificial intelligence agent should ideally be able to execute instructions in the same format. Past approaches have often involved separate modules, using a semantic parser to translate natural language to commands interpretable by the agent\citep{semanticparser2}, and a separately trained policy. In contrast, \citet{gatedattention} train an end-to-end reinforcement learning architecture to execute natural language instructions in a 3D visual environment. One of the primary challenges in grounding natural language for a visual agent is constructing an appropriate reward: how do we know when the agent has successfully carried out the instruction? 

The authors address this challenge by limiting the instructions to simple navigational commands, such as "go to the red torch". For such instructions, it is straightforward to generate a reward function by identifying the correct object from the description and adding a positive reward for navigating to that object, and a negative reward for navigating to any other object. 

For complicated instructions, generating a reward function is more difficult. A first obvious approach is to generate random environments and manually collect sentences and create the appropriate reward, but doing so would be very costly. Moreover, for any new environment of interest, rewards would have to be created anew. 

Our project contributes to reinforcement learning of natural language instructions by: 1) constructing a dataset that can be used to construct rewards for natural language navigation instructions across multiple environments, and 2) using this data to train an agent to carry out more complicated navigational instructions using the same environment and a similar architecture proposed in \citet{gatedattention}. We view the fact that the data is not tied to any particular environment as being particularly important, as this implies that the cost of future improvements to the dataset can be amortized over users of multiple environments, which allows for the collection and generation of more and better data.

We exploit the fact that natural language navigational instructions involve abstract descriptions of an environment, rather than being tied to a specific visual setting. In particular, we adapt the CLEVR visual question answering dataset and methodology \citep{clevr} to generate triplets of \textit{scene graphs}, which describe a set of objects, attributes, and spatial relationships between those objects, natural language instructions that uniquely identify an object in the scene graph, and the identity of the described object. We then use a mapping to generate an appropriate environment and reward from the scene graph for a specific setting, in our case, VizDoom. 

\section*{Related Work}
\textbf{Grounding task oriented instructions.} 
\cite{gatedattention} is the most important related work for our project, as we focus on an extension to the problem considered in that work. \cite{gatedattention} was the first work to perform reinforcement learning for language grounding in a 3D setting, specifically ViZDoom. We reuse their architecture and environment, which are described in detail later in the report. 


 \cite{stanfordinstructions} use natural language instructions to specify intermediate objectives, improving performance on Montezuma's revenge, which is an environment that requires longer-term planning. They hand-label frames to provide supervision for a neural network that takes a frame as an input and determines whether an instruction has been satisfied. With our approach we hope to eliminate the need to label vision frames and work with pre-labeled abstract representations instead. 
 
\cite{listenattendwalk} train a neural network to map natural language instructions to action sequences in a fully supervised setting. While we also use supervised data, our supervised data is sparse and only informs the rewards that the agent receives rather than specifying a full sequence of actions.

There is a large body of literature on using supervised data to inform agent policies, for instance on imitation learning. However, to our knowledge we are the first to focus on creating environment-agnostic supervision for language grounding. 

\textbf{Curriculum Learning}. In order to train the agent  to carry out more complex instructions in harder settings, we employ curriculum learning, where the agent is first trained on easier problems before moving on to harder problems. \cite{gatedattention} use a form of curriculum learning where they start training in an easy environment and move on to more difficult environments. In contrast, \cite{hill2017understanding} emphasize the importance of curriculum learning for learning language groundings. In particular, they gradually increase the vocabulary of the agent, expanding the vocabulary only when the previous vocabulary has been successfully grasped. In our work we employ both flavors of curriculum learning, moving on to environments with more objects and more complex instructions over time.

\section{Methods}

\subsection{Data Generation}

In order for language grounding data to transfer between environments, it needs to consist of matched sets of instructions and environments, where the environment data is not visual but contains information on all the objects in the environment. We found that an existing visual question-answering dataset, CLEVR \citep{clevr}, comes close to meeting these requirements. The CLEVR dataset contains 70000 scene graphs, which consist of a list of objects, the properties of those objects, namely the locations, size, color, material, and shape, and the spatial relationship between each pair of objects (to the left of, in front of, etc). They generate questions from these scenes using functional program templates, which are made up of a composition of a query and a set of filters, alongside text templates. 

An example of a CLEVR question might be "What is the color of the cylinder to the left of the red sphere?", created by composing (query color, filter shape, filter spacial relationship, filter color, filter shape). We have adapted these functional program and text templates to generate navigational instructions instead, by removing the query, selecting the filters to identify a unique object, and generating an instruction to travel to that object. One possible instruction corresponding to the CLEVR example question might be "Go to the cylinder to the left of the red sphere".

For this project we only train the model on instructions to this type of "Go to [object with properties] [spatial relationship] [object with properties]", but we have implemented templates for a wider variety of more complex instructions which can be used for training. To compare with our baseline, the instructions in \citet{gatedattention} are of the form "Go to [object with properties]".

\subsection{Environment Mapping}

We use the environment from \cite{gatedattention} for task-oriented language grounding, in which the agent can execute a natural language instruction and obtain a positive reward on successful completion of the task. The environment is built on top of the ViZDoom API \cite{vizdoom}, based on Doom, a classic first person shooting game.   It provides the raw visual information from a first-person perspective at every timestep. Each scenario in the environment comprises of an agent and a list of objects (a subset of ViZDoom objects) - one correct and rest incorrect in a customized map. The agent can interact with the environment by performing navigational actions turn left, turn right, move forward, and no action.

We map the CLEVR scene graphs to the VizDoom environment. CLEVR objects have a shape (cube, sphere, or cylinder), a size (small or large), are made from a material (metal or rubber) and are one of eight colors. Object materials do not map naturally to the Doom environment, so we use the attributes color, shape and size, leading to a total of 48 unique objects (8 colors, 2 shapes, 3 sizes) in our Doom environment. With 4 possible spatial relationships, this means there are $48 \cdot 48 \cdot 4 = 9216$ possible unique complex instructions of the type we are considering. 
\par The sphere, cube and cylinder shapes are mapped to the visually distinct doom objects column, skull and torch. Colors in CLEVR are mapped to the same color in Doom. Coordinates in the scene graph are mapped to Doom coordinates through a linear transformation that preserves relative distance between objects. 
$$ x_{doom} = \frac{x_{scene}(x_{doom}^{max} - x_{doom}^{min}) + (x_{scene}^{max}x_{doom}^{min} - x_{scene}^{min}x_{doom}^{max})}{x_{scene}^{max} - x_{scene}^{min} }$$
$$ y_{doom} = \frac{y_{scene}(y_{doom}^{max} - y_{doom}^{min}) + (y_{scene}^{max}y_{doom}^{min} - y_{scene}^{min}y_{doom}^{max})}{y_{scene}^{max} - y_{scene}^{min} }$$

\par The CLEVR scene graphs are generated from different view points with the camera looking at the objects. To correctly map the CLEVR scenes to our environment, we changed the orientation of the agent such that it could view the scene in the same manner as the camera. This was important for us as we have instructions like "Go to the small object to the LEFT of the blue torch". Such instructions would be meaningless if the scene was not viewed from the correct direction at the start of the episode and the agent would be unable to differentiate between left and right, which was a crucial part of our extension. 

\par
\begin{figure}[H]
\centering
\begin{subfigure}{.5\textwidth}
  \centering
  \fbox{\includegraphics[height=.5\linewidth]{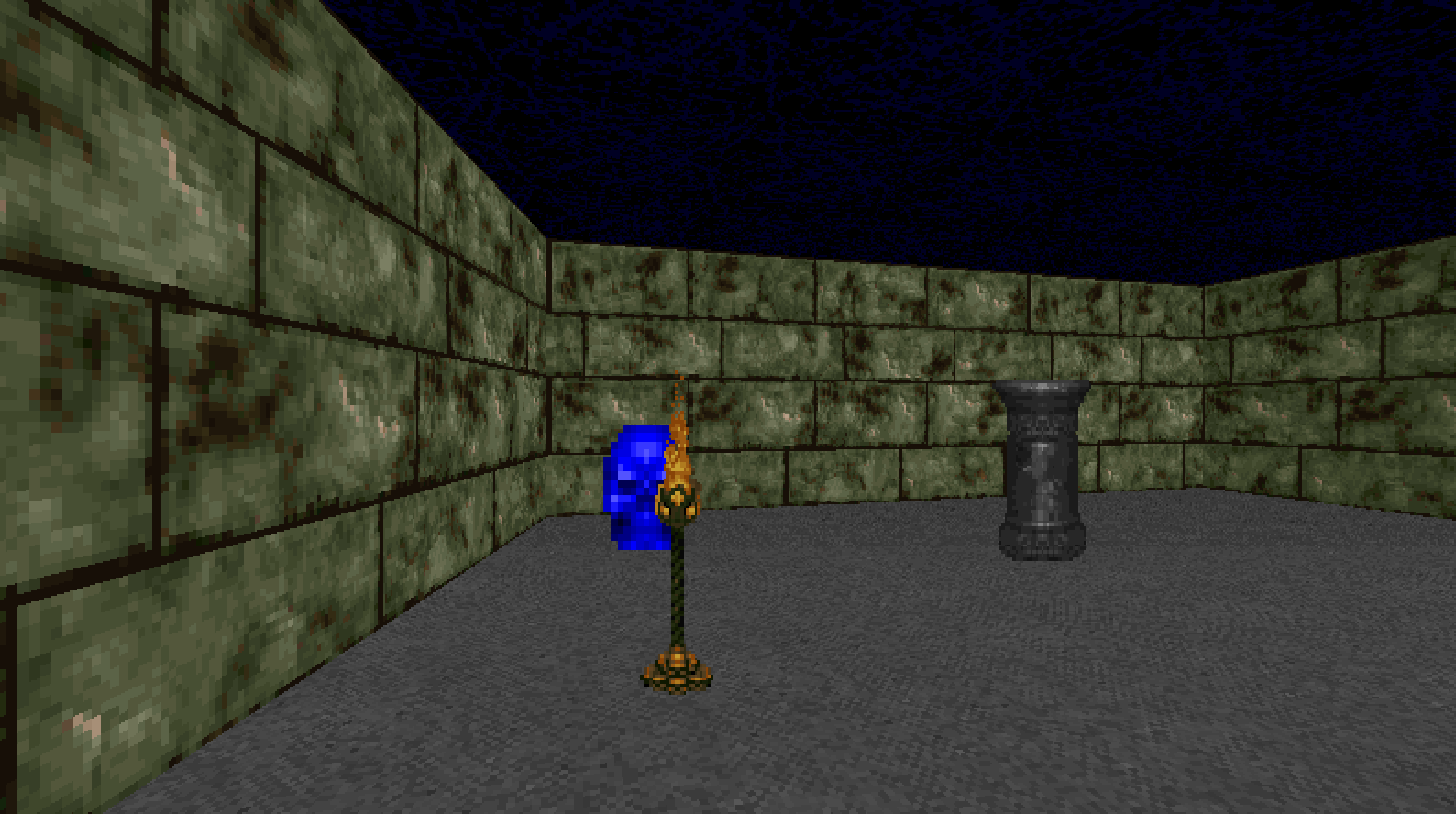}}
  \caption{Instruction: Go to the large gray object to the \\ right of the yellow torch}
  \label{fig:sub1}
\end{subfigure}%
\begin{subfigure}{.5\textwidth}
  \centering
  \fbox{\includegraphics[height=.5\linewidth]{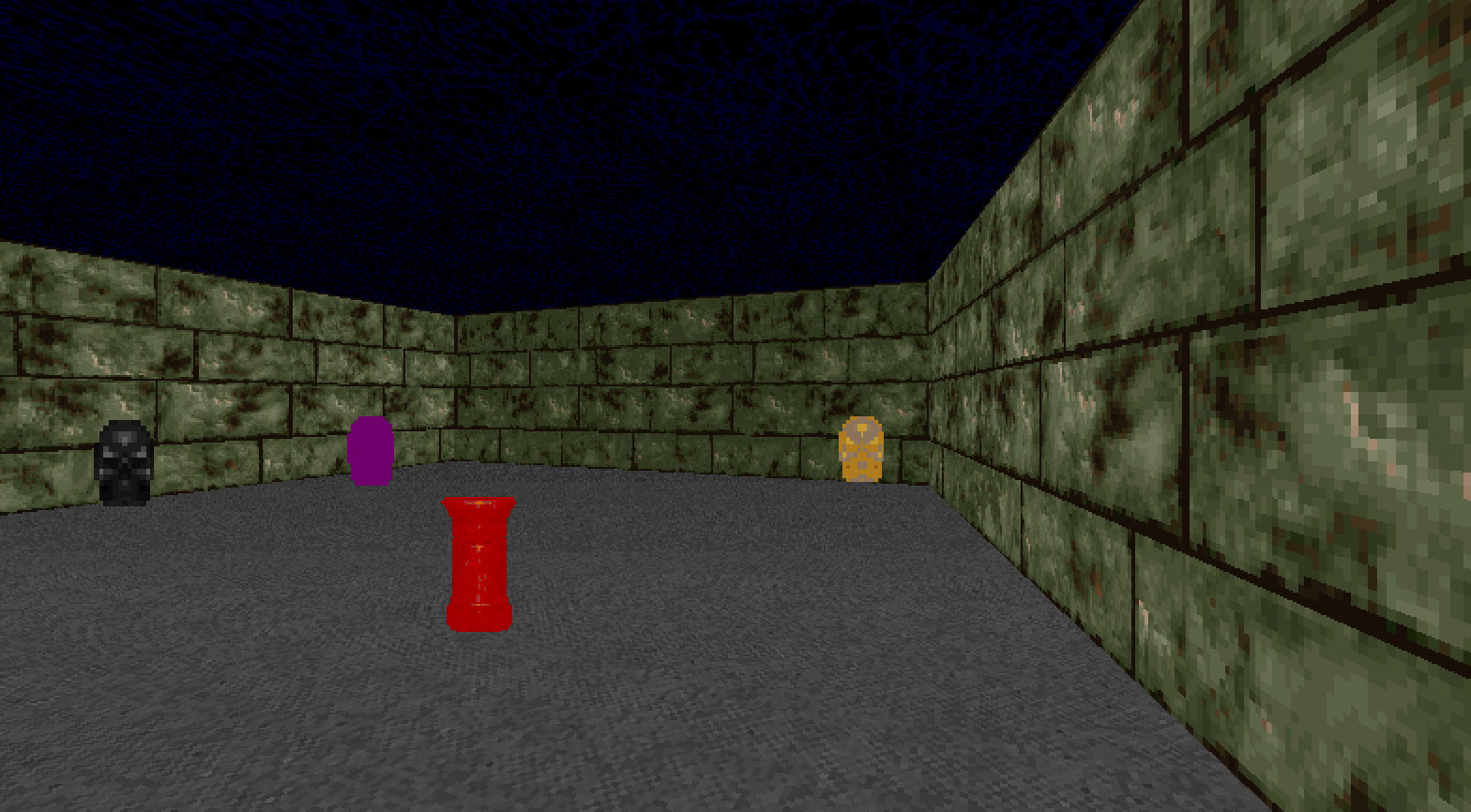}}
  \caption{Instruction : Go to the red column in front of the purple skull}
  \label{fig:sub2}
\end{subfigure}
\caption{CLEVR scenes mapped to Doom and corresponding instructions}
\label{fig:env1}
\end{figure}

 The instruction dataset consists of triplets of the scene ID, the instruction, and the identity of the target object. The environment for any instruction is created by loading the scene graph corresponding to the scene ID and spawning objects accordingly. Figure \ref{fig:env1} shows examples of mapped instruction-environment pairs. 

\subsection{Network Architecture}

\begin{figure}[H]
\centering      
\fbox{\includegraphics[width=\textwidth]{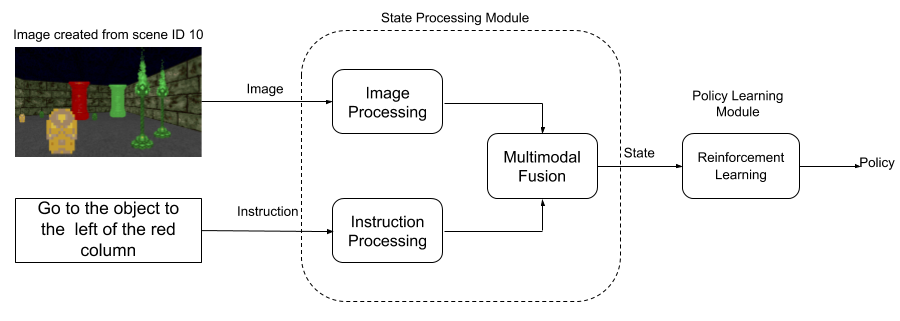}}
\caption{Model Architecture }
\label{fig:arch}
\end{figure}

For our baseline implementation, we followed the architecture proposed by \citep{gatedattention} which comprises of a state processing module and a policy learning module, as shown in \ref{fig:arch}. The state processing module takes as input the current state, which in our setting is defined as a combination of a natural language instruction and a raw pixel-level image describing the environment, and creates a joint representation for them. This joint representation is used by the policy learner to predict the optimal action to take at that timestep. 

\textbf{State Processing Module:} consists of a convolutional network to process the image, a Gated Recurrent Unit (GRU) network to process the instruction and a multimodal fusion unit, \textit{Gated Attention}, to combine the representations of the instruction and the image. In the Gated-Attention unit, the instruction embedding is passed through a fully-connected linear layer with a sigmoid activation. The output of this linear layer is called the attention vector and the output dimension is equal to the number of feature maps in the output of the convolutional network.

\textbf{Policy Learning Module:} applies the Asynchronous Advantage Actor-Critic (A3C) algorithm which uses a deep neural network to learn the policy and value functions and runs multiple parallel threads to update the network parameters. They also use entropy regularization for improved exploration and the Generalized Advantage Estimator to reduce the variance of the policy gradient updates. The policy learning module consists of an LSTM layer, followed by fully connected layers to estimate the policy function as well as the value function. The LSTM layer is introduced so that the agent can have some memory of previous states. If the agent is looking for an object, it is important to retain information about where it has already looked. Moreover, complicated instructions such as "Go to the object to the left of the red column" are only valid with respect to a particular frame of reference, so the agent needs to remember either that frame of reference or the object it has identified.

\section{Experiments}

We perform 3 sets of experiments: 1) apply the architecture from \cite{gatedattention} to scene graphs with complex instructions directly,  2) employ a curriculum learning approach in which we first train the model using simple instructions and few objects, and gradually increase the proportion of complex instructions and the number of objects, and 3) experiment with different architectures and reward structures. In addition we also show the performance of the baseline in \cite{gatedattention} (without CLEVR data) and the midterm report baseline for comparison. 

The training data consists of 14000 CLEVR scenes with 5 complex and 5 simple generated instructions each. During training, the objects are spawned according to the spatial relations mentioned in scene graph. The coordinates in scene graph were linearly mapped to the doom environment coordinates. One training episode begins with randomly selecting an instruction followed by environment generation. The episode terminates if the agent reaches the target or times out (T = 30). The models are evaluated based on the accuracy to reach the target before episode terminates. 

\subsection{Training details}
The reward structure of our environment is the same as \cite{gatedattention}, in which -0.2 is awarded for reaching a wrong object, 0 for not reaching any object before timing out and 1 for the correct target. 

\par The input to the neural network is the instruction and an RGB image of size $3\times 300 \times 168$. The first layer convolves the image with 128 filters of $8\times 8$ kernel size with stride 4, followed by 64 filters of 4x4 kernel size with stride 2 and another 64 filters of $4\times 4$ kernel size with stride 2. The input instruction is encoded through a Gated Recurrent Unit (GRU)\cite{chung2014empirical} of size 256.
\par For reinforcement learning, we run experiments with A3C algorithm. The policy learning module has a linear layer of size 256 followed by an LSTM layer of size 256 which encodes the history of state observations. The LSTM layer's output is fully-connected to a single neuron to predict the value function as well as three other neurons to predict the policy function. All the network parameters are shared for predicting both the value function and the policy function except the final fully connected layer. All the convolutional layers and fully-connected linear layers have ReLu activations \cite{nair2010rectified}. The A3C model was trained using Stochastic Gradient Descent (SGD) with a learning rate of 0.001. We used a discount factor of 0.99 for calculating expected rewards. 
We use mean-squared loss between the estimated value function and discounted sum of rewards for training with respect to the value function, and the policy gradient loss using for training with respect to the policy function. The model was trained on an Amazon C5 instance, running 36 parallel threads over 72 cores. 

For the curriculum learning experiment we followed the following training sequence: 

\begin{enumerate}
\item 3 objects with only simple instructions
\item 3 objects with proportion 0.1 complex instructions
\item 3 objects with proportion 0.5 complex instructions
\item 3 objects with proportion 0.75 complex instructions
\item 5 objects with proportion 0.5 complex instructions
\end{enumerate}


\section{Results and discussion}

\begin{figure}[H]
\centering
  \centering
  \fbox{\includegraphics[width=.5\linewidth]{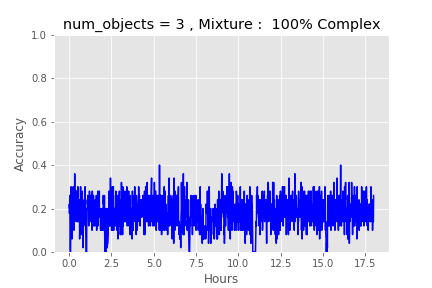}}
  \label{fig:complexonly}
\caption{Training only on complex instructions}
\label{fig:res4}
\end{figure}

The results for the first experiment, training only on complex instructions, are shown in Figure \ref{fig:res4}. The model fails to learn anything. We hypothesize that the model is being asked to learn too many things at the same time. In the case of the simple instructions, the model has to learn representations of objects and properties of those objects as well as a policy for navigating towards those objects. For the more complex instructions, the model has to learn spatial relationships at the same time, a difficult task.

\begin{figure}[H]
\centering
\begin{subfigure}{.5\textwidth}
  \centering
  \fbox{\includegraphics[width=.8\linewidth]{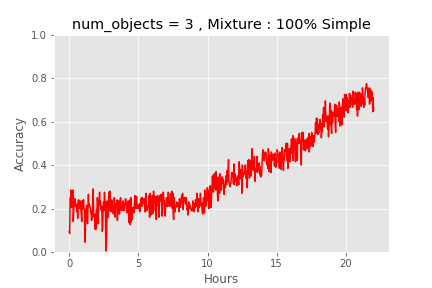}}
  \caption{3 Objects, only simple instructions}
  \label{fig:cl1}
\end{subfigure}%
\begin{subfigure}{.5\textwidth}
  \centering
  \fbox{\includegraphics[width=.8\linewidth]{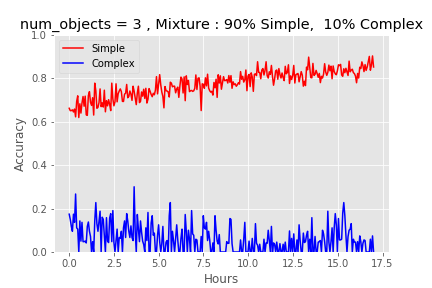}}
  \caption{3 objects, 0.1 proportion complex instructions}
  \label{fig:cl2}
\end{subfigure}
\begin{subfigure}{.5\textwidth}
  \centering
  \fbox{\includegraphics[width=.8\linewidth]{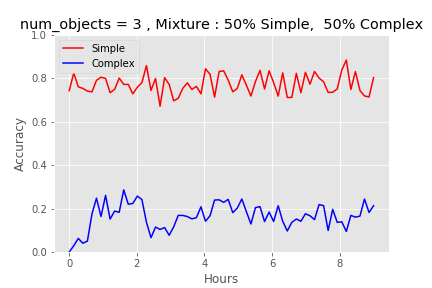}}
  \caption{3 objects, 0.5 proportion complex instructions}
  \label{fig:cl3}
\end{subfigure}%
\begin{subfigure}{.5\textwidth}
  \centering
  \fbox{\includegraphics[width=.8\linewidth]{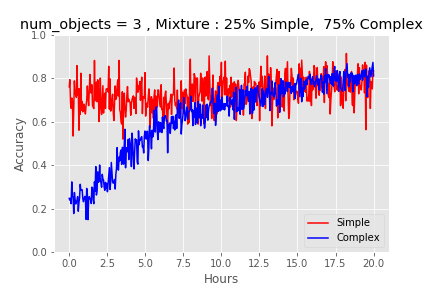}}
  \caption{3 objects, 0.75 proportion complex instructions}
  \label{fig:cl4}
\end{subfigure}\\
\begin{subfigure}{.5\textwidth}
  \centering
  \fbox{\includegraphics[width=.8\linewidth]{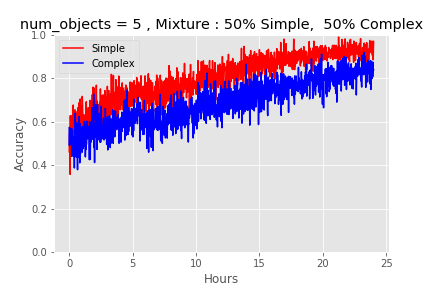}}
  \caption{5 objects, 0.5 proportion complex instructions }
  \label{fig:cl5}
\end{subfigure}%
\caption{Curriculum Learning}
\label{fig:curriculum}
\end{figure}

We predicted that training on simple instructions first in the curriculum learning experiment would allow for this training to proceed in stages, so that the model can first learn a representation of objects and only then learn to represent spatial relationships. Table  \ref{tab:curriculum} and Figure \ref{fig:curriculum} show the results of the curriculum learning experiment. Indeed, we find that training the model on simple instructions first leads accuracy on complex instructions to improve after a much smaller number of iterations then we ran for complex instructions only.

\begin{table}[H]
\centering

\caption{Final average accuracy for curriculum learning models over 300 instructions}
\begin{tabular}{l|c|c}  \hline
Model & Simple accuracy & Complex accuracy \\  \hline 
3 Objects, only simple &0.66 & n/a \\
3 Objects, 0.1 complex &0.81 & 0.08 \\
3 Objects, 0.5 complex & 0.75 & 0.30 \\
3 Objects, 0.75 complex & 0.76&0.79 \\
5 Objects, 0.5 complex &0.95 &0.86 \\ \hline 
\end{tabular}
\label{tab:curriculum}   
\end{table}

\begin{figure}
\centering
\begin{subfigure}{.5\textwidth}
  \centering
  \fbox{\includegraphics[width=.8\linewidth]{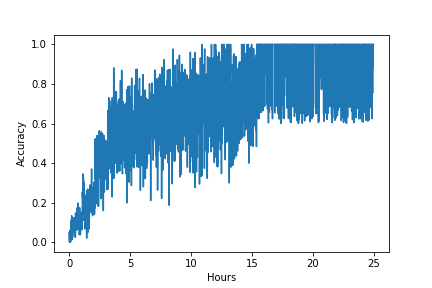}}
  \caption{Results of reproduction \citep{gatedattention}}
  \label{fig:baseline}
\end{subfigure}%
\begin{subfigure}{.5\textwidth}
  \centering
  \fbox{\includegraphics[width=.8\linewidth]{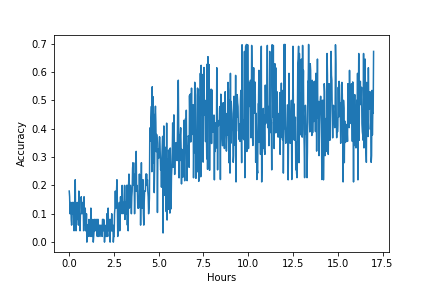}}
  \caption{Results of model with complex instructions}
  \label{fig:sub2}
\end{subfigure}
\caption{Midway report baseline performance of complex instructions}
\label{fig:midway}
\end{figure}

We considerably improved on the accuracy from the midway report (Figure \ref{fig:midway}) and come close to the baseline accuracy from \cite{gatedattention} (see Figure \ref{fig:baseline}). Accuracy was still increasing when we stopped training due to computational constraints, so it is possible that both sets of instructions would have reached the level of accuracy of the baseline. We do observe that the accuracy of the complex instructions still lags behind by approximately 0.1.

\subsection{Error Analysis}
We visually inspected a number of mistakes to determine the most common errors. The errors can be grouped roughly into two categories which occur at similar frequencies: navigation errors, in which the agent has identified the correct object but fails to navigate to that object, and identification errors, in which the agent fails to identify the correct object. Navigation errors occur either when the agent barely grazes an object at the edge of its vision, or when the target object is obstructed by other objects so that the correct path is very circuitous. The grazing issue might be avoided by improving the memory component of the model, such that the agent remembers there is an object close beside it. Planning a complicated route is a more difficult task for the agent to learn and might not be possible with the current architecture.  

Identification errors occur either when the correct object is not in view and the agent rotates in the wrong direction, or in the case of the complex instructions, when the agent becomes confused between the target object and the object that is used as a reference. This is more likely to occur when the instruction contains no description of the target object. For example, for the instruction `go to the object to the left of the red cylinder' the agent sometimes navigates towards the red cylinder. Another type of identification error occurs when the agent moves to a location or orientation on the map where the original instruction is no longer valid. Instructions such as `Go to the torch on the left side of the skull' are only necessarily correct relative to the frame of reference at the start of the episode - the agent could travel or rotate in such a way that the instruction is no longer valid given its current positioning. Therefore the agent has to keep an updated representation of either the instruction or the state. The LSTM allows it to do this in principle, but it appears to be difficult to learn and practice

\subsection{Architecture and Training Experiments}

\begin{figure}
\centering
\begin{subfigure}{.5\textwidth}
  \centering
  \fbox{\includegraphics[width=.8\linewidth]{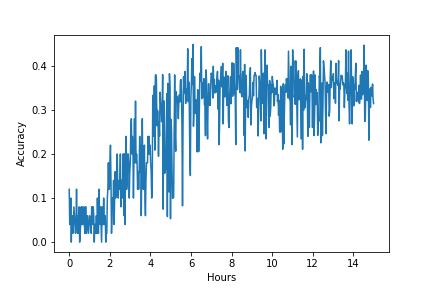}}
  \caption{Self Attention(\citep{lin2017structured}) Instruction Processing Architecture}
  \label{fig:satya1}
\end{subfigure}%
\begin{subfigure}{.5\textwidth}
  \centering
  \fbox{\includegraphics[width=.8\linewidth]{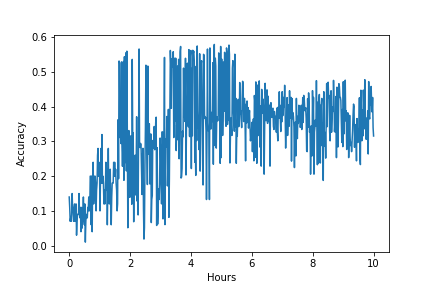}}
  \caption{Modified reward function }
  \label{fig:satya2}
\end{subfigure}
\caption{Experiment results for Model Tuning and Modified Reward Structure }
\label{fig:satya}
\end{figure}

In order to improve performance of complex instructions, we also experimented with different instruction processing models, such as encoding the instructions with self-attention rather than a GRU network. \cite{lin2017structured} report better performance of the self attention network for multiple tasks. In our case, it performed worse than the original model (see Figure \ref{fig:satya1}), possibly due to the fact that our sentences are not too long with not many clauses, as it seems to work worse for single-clause sentences. 

\par 
Also, we experimented with an alternate denser reward function as we consider the possibility that the reward was too sparse for the agent to learn. Hence, we changed the reward function such that it gets -0.1 for every step until it hits any object, +10 for hitting the target object and -5 for hitting the wrong ones.  We noticed that this improved convergence at the expense of accuracy (Figure \ref{fig:satya2}). All these experiments were done for 10\% complex and 90\% simple instructions.

\section{Conclusion and Future Work}
In this project, we adapted the CLEVR data set to generate a set of complicated natural language instructions, and mapped CLEVR scenes to VizDoom environments in order to provide supervision for a reinforcement learning agent to learn to carry out those instructions. We used the setup and architecture from  \cite{gatedattention}, and employed curriculum learning to overcome the difficulty of the more complex instructions, starting with the simpler original instructions from \cite{gatedattention} and adding in more complex instructions gradually. Our final trained agent carries out both types of instructions with high accuracy.

We view our contribution as consisting of 1) demonstrating that the gated fusion architecture is capable of learning to execute more complex navigational instructions instructions and 2) providing a proof of concept for a general natural language instruction dataset for visual reinforcement learning agents, separated from any particular environment. The fact that the instructions can be separated from any particular environment implies that the cost of creating a dataset can be amortized over the different environments in which it is used, allowing for more and increasingly complex data. In the future, we intend to capitalize on this by constructing a dataset with a larger variety of complex and compound instructions and demonstrate its use in different reinforcement learning environments.
\medskip

\bibliography{nips_2016}
\bibliographystyle{acl_natbib}

\end{document}